\documentclass[copyright,creativecommons]{eptcs}
\providecommand{\event}{HAS 2013} 
\usepackage{breakurl}             
\usepackage{graphicx}
\usepackage{color,amsfonts}

\title{A stochastic hybrid model of a biological filter}
\author{
Andrea Ocone 
\institute{School of Informatics\\University of Edinburgh\\Edinburgh, Scotland}
\email{a.ocone@ed.ac.uk}
\and
Guido Sanguinetti
\institute{School of Informatics\\University of Edinburgh\\Edinburgh, Scotland}
\email{G.Sanguinetti@ed.ac.uk}
}
\def\titlerunning{Hybrid biological filters}
\def\authorrunning{A. Ocone \& G. Sanguinetti}
\begin{document}
\maketitle

\begin{abstract}
We present a hybrid model of a biological filter, a genetic circuit which removes fast fluctuations in the cell's internal representation of the extra cellular environment. The model takes the classic feed-forward loop (FFL) motif and represents it as a network of continuous protein concentrations and binary, unobserved  gene promoter states. We address the problem of statistical inference and parameter learning for this class of models from partial, discrete time observations.  We show that the hybrid representation leads to an efficient algorithm for approximate statistical inference in this circuit, and show its effectiveness on a simulated data set.
\end{abstract}

\section{Introduction}

\begin{figure}[htpb]
\begin{center}
\includegraphics[width=4cm]{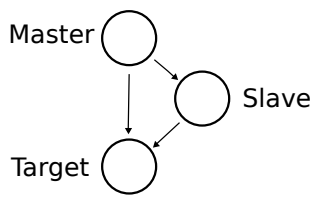}
\caption{Feed-forward loop structure.}\label{fig001}
\end{center}
\end{figure}

Organisms exist in a constantly changing and noisy environment. In order to carry out many fundamental functions, cells need to represent internally changes in environmental conditions, and to process what are effectively highly noisy signals. In many simple organisms, this internal representation is achieved via the chemical modification of a specific class of proteins, transcription factors (TFs), which are able to bind DNA and to modulate the expression of downstream genes \cite{Ptashne:genes02}. In many cases, the proteins which are the products of these downstream genes will then exert a chemical feedback on the initial stimulus (e.g. by breaking down a nutrient). 
While this is a simple and effective approach to respond to environmental stimuli, the energetic costs of protein production are high, and it would be in many cases undesirable to respond to signals which are not present for a sufficiently long time. As a consequence, cells have evolved regulatory structures which are able to filter out rapid fluctuations \cite{Alon:introduction06}. A prototypical, and very common, example is the feed-forward loop (FFL), shown in Figure~\ref{fig001}: here the final protein product is activated by two TFs, a {\it master} TF and a {\it slave} TF, which is regulated by the master. Therefore, in order for the final protein to be produced, the input signal must be present for a sufficient time to enable the production of the slave TF in sufficient quantities.
Transcription (and translation) in a single cell is essentially a stochastic process \cite{Elowitz:stochastic02,Swain:intrinsic02}, with mRNAs produced in individual units as a result of the change in occupancy of the promoter. Nevertheless, a discrete state description of the system may be problematic from a computational point of view (when the number of molecular species is large), so that it is often convenient to approximate the discrete number of mRNAs with a continuous random variable (its concentration). This mesoscopic approximation can be shown to be asymptotically correct in the large number limit \cite{Kampen:stochastic81}, and has been recently used in a formal modelling context~\cite{Bortolussi:fluid12}. Here, we couple such a mesoscopic approximation with the binary state of the gene promoters, naturally obtaining a description of the system as a hybrid system. This hybrid system is essentially a stochastic version of the well known ON/ OFF model of gene expression \cite{Ptashne:genes02}.

Assessing how well a model represents a real biological system is a non-trivial challenge. In general, model parameters (transcription/ decay rates, binding affinities, etc...) are only approximately known. Furthermore, a model such as the hybrid model we propose for transcription contains unobserved species (in this case, the promoter's state), whose dynamics constitute an important part of the model. Bayesian statistics offers a principled solution to both of these problems: uncertainty over parameters can be incorporated through prior distributions, and the dynamics of the unobserved species can be reconstructed {\it a posteriori}, providing important detail on how external signals are handled by the circuit in specific cases. Statistical inference in the ON/ OFF and related models of gene expression has been addressed very recently both for deterministic \cite{Sanguinetti:switching09,Opper:learning10,Ocone:reconstructing11} and stochastic systems \cite{Opper:approximate10}. Here we present an extension and application of the stochastic framework in \cite{Opper:approximate10} to the feed-forward loop architecture. The primary novelty lies in the handling of the slave TF, in particular how its promoter state depends on the master TF protein concentration. As discussed later, this is in marked contrast to the approach taken in \cite{Ocone:reconstructing11} for deterministic systems, and leads to substantial gains both computationally and in terms of identifiability. 

The rest of the paper is organised as follows. We start off by presenting the general framework we adopt, and by reviewing the related works made over the last few years. We then describe the specific model we consider in this contribution, and the associated approximate inference algorithm. This algorithm is implemented and tested on a simulated data set. We conclude by discussing the merits and limits of our work in the broader context of dynamic modelling, and discussing future work.

\section{Stochastic ON/OFF models of gene expression}

Our hybrid models describe fundamental gene regulation mechanisms such as transcriptional activation and translation with a system of differential equations. We represent two main physical entities: discrete binary promoter states $\mu$, which can be occupied ($\mu=1$) or not ($\mu=0$), and continuous protein states $x$. Figure~\ref{fig002} shows a graphical representation of these entities: discrete variables are represented as squares and continuous variables as circles. Promoter and protein states are linked together through the following transcription-translation stochastic differential equation (SDE) model\footnote{In Equation~(\ref{proteinModel}) we are considering transcription and translation mechanisms combined in a single reaction, by assuming that the protein states are proportional to corresponding mRNA levels. }:
\begin{equation}\label{proteinModel}
\mathrm{d}x(t) = \Big(b - \lambda x + A\mu(t)\Big)\mathrm{d}t + \sigma \mathrm{d}w(t)\,.
\end{equation}
Here $b$, $\lambda$ and $A$ are kinetics parameters: $b$ is a basal transcriptional-translational rate; $\lambda$ is the protein decay constant (proportional to the inverse of the protein half life); $A$ represents an increment (or decrement) of the transcriptional-translational rate occurring when the promoter is occupied. If the promoter is occupied ($\mu=1$), the transcription-translation rate is then given by $A+b$; otherwise, when $\mu=0$, the transcription-translation rate is simply given by $b$. Therefore there are two distinct levels of transcription-translation rate, corresponding to the discrete ON/OFF state of the promoter \cite{Ptashne:genes02}. While $b$ and $\lambda$ are constrained to be positive, the parameter $A$ can be positive or negative to model activation or repression, respectively. The final term in Equation~(\ref{proteinModel}), $\sigma dw(t)$, takes into account of the stochasticity of the transcription-translation mechanism: $w(t)$ is a Wiener process and $\sigma$ represents the system noise variance. Note that when $\mu=0$, Equation~(\ref{proteinModel}) reduces to a simple Ornstein-Uhlenbeck process. In this case the protein state is described by a Gaussian distribution with mean
\begin{equation}
\mathbb{E}[ x(t) | \mu(t) = 0 ] = x(0) \mathrm{e}^{-\lambda t} +\frac{b}{\lambda} \left( 1 - \mathrm{e}^{-\lambda t} \right)\,,
\end{equation}
where $x(0)$ is the protein state at time $t=0$.

To model the discrete promoter states $\mu$ we use a two-state Markov jump process, also known as \emph{telegraph process}. This represents a stochastic process that switches between two discrete states (in our case $\mu=0$ and $\mu=1$) and whose single time marginal probability is described by the chemical master equation
\begin{equation}\label{master}
\begin{array}{rcl}
\displaystyle \frac{\mathrm{d}}{\mathrm{d}t} p_{\mu}(1,t) & = & f_{+}p_{\mu}(0,t)-f_{-}p_{\mu}(1,t)\,,\\\\
\displaystyle \frac{\mathrm{d}}{\mathrm{d}t} p_{\mu}(0,t) & = & f_{-}p_{\mu}(1,t)-f_{+}p_{\mu}(0,t)\,.
\end{array}
\end{equation}
which, using the fact that probabilities sum to $1$ ($p_{\mu}(1,t)+p_{\mu}(0,t)=1$), can be reduced to
\begin{equation}
\frac{\mathrm{d}}{\mathrm{d}t} p_{\mu}(1,t) = f_{+} - (f_{+}+f_{-})p_{\mu}(1,t)\,.
\end{equation}
The quantity $p_{\mu}(1,t)$ (or $p_{\mu}(0,t)$) represents the probability of the promoter state $\mu$ to be $1$ at time $t$ (or $0$ at time $t$). The parameters $f_{+}$ and $f_{-}$, known as switching rates, are probabilities per unit time to switch from state $0$ to state $1$ and vice versa, respectively.

By using a telegraph process, the behaviour of the promoter states is described by discrete jumps that are much faster compared to the continuous evolution of the protein states. This assumption is reasonably motivated by the fact that the promoter binding reaction is much faster compared to the time needed for the transcription-translation process.

The combination of a protein $x$ and its correspondent promoter $\mu$ represents the fundamental unit of our hybrid models. In order to build a network motif such as the feed-forward loop, we need to combine together a number of these units. Therefore, we also model the dependence of $\mu$ on the state of an upstream protein $x'$ (which is a TF for the promoter $\mu$). 

To link together promoter states $\mu$ and upstream continuous protein states $x'$ we use the following relations for the switching rates of the promoter
\begin{eqnarray}
 f_+ &=& k_p \exp \left( k_e x' \right) \,,\label{prior1}\\
 f_- &=& k_m \,,\label{prior2}
\end{eqnarray}
where $k_p$, $k_e$ and $k_m$ are additional parameters. By writing the switching rate $f_+$ as a function of $x'$ (and consequently of time $t$), we obtain that the probability of the promoter to switch from state $0$ to state $1$ depends on the $x'$ concentration. On the other hand $f_-$ is kept constant to the parameter value $k_m$.

Note that modelling the promoter states with the master Equation~(\ref{master}) and the switching rates as in Equation~(\ref{prior1}) and~(\ref{prior2}), we are assuming a saturation effect of the promoter $\mu$ by the transcription factor concentration $x'$. From the master equation, we obtain that the steady state probability of the promoter being ON $p_{SS}(\mu = 1|x')$ corresponds to a Hill type dependence upon the (exponential of the) upstream protein concentration $x'$:
\begin{equation}
p_{SS}(\mu = 1|x') = \frac{f_+}{f_-+f_+} = \displaystyle \frac{\exp (k_e x')}{\frac{k_m}{k_p} + \exp (k_e x')} \,.
\end{equation}

\section{Related work}
The idea of using models with latent variables for transcriptional regulation has a long history in computational biology: the first attempts to integrate microarray data with the architecture of the regulatory network in order to infer unobserved transcription factor behaviours date to the early days of ChIP-on-chip technology \cite{Liao:network03,Sanguinetti:probabilistic06}. The idea to use a differential equation model for transcription within a statistical context is more recent. In early models \cite{Barenco:ranked06,Lawrence:modelling06}, the prior assumptions on TF profiles (piecewise constant or Gaussian) were relatively weak and not particularly linked to accepted biological models of transcription factor behaviour. Sanguinetti et al \cite{Sanguinetti:switching09} recasted the deterministic ON/OFF model of gene expression \cite{Ptashne:genes02} in statistical terms, deriving algorithms for both exact and variational inference. This approach proved fruitful and enabled generalisations to more complex models of regulation, including stochastic models \cite{Opper:approximate10}, combinatorial regulation \cite{Opper:learning10}, hierarchical networks \cite{Ocone:reconstructing11} and networks of arbitrary topology~\cite{OconeMillarSanguinetti}. Stochastic hybrid systems of this kind have been studied quite intensely in earlier works~\cite{Pola:stochastic03}.

While the FFL network has already been studied in \cite{Ocone:reconstructing11}, that study only considered deterministic systems, and modelled the saturating effects of regulation by the slave TF through the use of a Heaviside step function. This non-differentiability introduced severe computational problems: in particular, identifiability of the saturation threshold was weak and could only be done by direct search, with considerable computational overheads. By contrast, here we adapt the stochastic approach of~\cite{OconeMillarSanguinetti} to the FFL topology, which avoids these computational issues by modelling directly the impact of the slave transcription factor on the switching rates of the target's promoter.

\section{The hybrid feed-forward loop model}

FFLs consist of three components: a master TF which regulates the transcriptional activity of both a slave gene and a target gene. In turn, the slave TF also regulates the target gene. The presence of three regulatory connections with two possible signs (activation or repression) gives rise to $2^3$ types of FFLs.

Among the $2^3$ types, the most recurrent is the one with all positive connections, with frequency of about $50\%$ and $40\%$ in yeast and E.coli, respectively~\cite{Alon:introduction06}. The reason is that this particular FFL can work as a filter against spurious fluctuations of biological signals. The mechanism is the following. The target protein is produced only if both master and slave TF are present. If a noisy high frequency signal causes the undesired production of master TF, then, before activating also the target gene, this signal is delayed by the production of slave TF. Therefore, at the level of the target gene, it is filtered out.
\begin{figure}[htpb]
\begin{center}
\includegraphics[width=5cm]{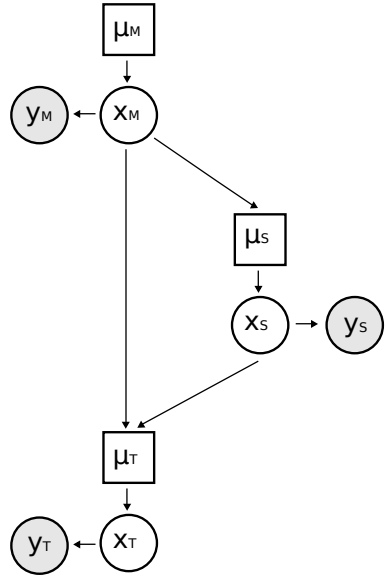}
\caption{Hybrid FFL model. Squares represent promoter states, empty circles represent protein states and grey circles represent observations.}\label{fig002}
\end{center}
\end{figure}

As showed in Figure~\ref{fig002}, our hybrid FFL model is composed of three units promoter-protein. If we assume that initially the promoter state $\mu_M$ is $0$, then in presence of an external stimulus, the master promoter $\mu_M$ changes its activity from $0$ to $1$, and starts the transcription-translation of master protein $x_M$. The master protein affects the switching rate of the slave promoter $f_{S+}$. Then the probability of the slave promoter to switch from state $0$ to state $1$ increases and transcription-activation of slave protein occurs. Together with the master protein, the slave protein changes the switching rate of the target promoter $f_{T+}$, which finally brings to transcription-translation of the target protein. If the external stimulus is too short (e.g. noisy fluctuations) then the transcription-translation of the target protein will not occur. Mathematically this is described by the following system of differential equations:
\begin{eqnarray}
\frac{\mathrm{d}}{\mathrm{d}t} p_{\mu_M}(1,t) &=& f_{M+} - (f_{M+}+f_{M-})p_{\mu_M}(1,t)\\
\mathrm{d}x_M(t) &=& \Big(b_M - \lambda_M x_M + A_M\mu_M(t)\Big)\mathrm{d}t + \sigma \mathrm{d}w(t)\label{FFL1}\\
\frac{\mathrm{d}}{\mathrm{d}t} p_{\mu_S}(1,t) &=& f_{S+} - (f_{S+}+f_{S-})p_{\mu_S}(1,t)\\
\mathrm{d}x_S(t) &=& \Big(b_S - \lambda_S x_S + A_S\mu_S(t)\Big)\mathrm{d}t + \sigma \mathrm{d}w(t)\\
\frac{\mathrm{d}}{\mathrm{d}t} p_{\mu_T}(1,t) &=& f_{T+} - (f_{T+}+f_{T-})p_{\mu_T}(1,t)\\
\mathrm{d}x_T(t) &=& \Big(b_T - \lambda_T x_T + A_T\mu_T(t)\Big)\mathrm{d}t + \sigma \mathrm{d}w(t)\,,\label{FFLend}
\end{eqnarray}
where the switching rates of the slave and target genes are
\begin{eqnarray}
 f_{S+} &=& k_p \exp \left( k_e x_M \right)\nonumber\\
 f_{S-} &=& k_m \nonumber\\
 f_{T+} &=& k_p \exp \left( k_e \frac{1}{2}\left( x_M + x_S \right) \right)\nonumber\\
 f_{T-} &=& k_m \nonumber\,.
\end{eqnarray}
The switching rate $f_{T+}$ is a function of the average of the protein states $x_M$ and $x_S$. This means that the probability to switch the target promoter state from $0$ to $1$ requires the presence of both master and slave proteins. If the signal on the master gene is too short, then when $x_S$ starts to be produced, the protein $x_M$ is already decaying. Therefore the probability $f_{T+}$ will not increase.

The number of parameters to estimate in our hybrid FFL model is $3\times 3$: the three kinetics parameters for each of the three gene. Instead, we do not estimate the parameters $k_p$, $k_e$ and $k_m$, that are fixed to arbitrarily chosen values.

The parameters are estimated from measurements of protein levels, which are usually available in the form of noisy time-series. Therefore we consider the protein measurements as discrete-time observations of the true continuous-time protein levels. By assuming the observations $y_{i\,gene}$ to be i.i.d. (the index ${gene}$ refers to $M$, $S$ and $T$), we have the following Gaussian likelihood noise model:
\begin{equation}\label{likel}
\mathcal{L} = \prod_{i=1}^N p(y_{i\,gene}|x_{gene}(t_i)) = \prod_{i=1}^N \mathcal{N}(y_{i\,gene}|x_{gene}(t_i),\sigma_{\mathrm{obs}}^2)\,,
\end{equation}
where $N$ is the total number of observations and $\sigma_{\mathrm{obs}}^2$ is the observation variance. Using corrupted data $D_{gene} = \{y_{1\,gene},y_{2\,gene},\ldots,y_{N\,gene}\}$, we are then interested in reconstructing the true protein states $x_{gene}$. In Figure~\ref{fig002}, observations are represented as shaded nodes.

On the other hand, observations of promoter states are usually not available, so we model $\mu_{gene}$ as unobserved (or latent) variables which have to be inferred as well from the data $D_{gene}$. An advantage of using a latent variable representation, is that the model becomes very flexible and so capable to capture highly nonlinear network dynamics.

\section{Approximate inference}

The process $x$ is not Markovian (Eq.~\ref{proteinModel}), since it depends also on the state of the telegraph process. But if we consider the joint process $[x,\mu]$, then this is Markovian. Given the Markovian nature of the joint process, we can use the forward-backward algorithm to solve exactly the inference problem. This exact inference solution is expensive from a computational perspective, because it involves the numerical solution of coupled partial differential equations to find the posterior distributions. Then we adopt a tractable solution to the inference problem, using an approximate inference framework.

The posterior distributions of the joint process $[x,\mu]$ given the noise observations $D$ is given by Bayes' rule
\begin{equation}
p(x,\mu|D) = \frac{1}{Z} p(x,\mu) \mathcal{L}\,,
\end{equation}
where $p(x,\mu)$ is the prior distribution, $\mathcal{L}$ is the Gaussian likelihood (Eq.~\ref{likel}) and $Z$ represents the marginal likelihood. To solve the inference problem (i.e. compute $p(x,\mu|D)$) we adopt a variational method. Variational methods are a family of deterministic approximations 
which are based on bounding properties of the marginal likelihood~\cite{Bishop:pattern06}. They essentially consist of two steps: the first is to transform the inference problem into an optimisation problem; the second is to look for approximate solutions to the optimisation problem.

The optimisation problem is defined by choosing a so called variational distribution $q(x,\mu)$, which can approximate our target distribution $p(x,\mu|D)$, and an objective function $D(q,p)$ to minimise. In order to obtain tractable computations we choose the relative entropy, also known as Kullback-Leibler (KL) divergence:
\begin{equation}
D(q,p) = \mathrm{KL}(q \| p) = \int q \log \frac{q}{p} \mathrm{d} q \,.
\end{equation}
The KL divergence $KL(q\|p)$ satisfies the property that it is always positive and becomes null if and only if $q = p$. The variational density $q(x,\mu)$ is chosen within a family of tractable distributions and is function of some variational parameters; therefore the optimisation problem reduces to find a set of values of the variational parameters for which the KL divergence is minimal.

To allow for tractable computations, we relax the optimisation problem by making some approximations. First, we use a mean-field approximation for the variational distribution~\cite{Opper:advanced01} which is essentially obtained by making assumptions about the structure of the variational distribution $q(x,\mu)$. In particular, we assume that the variational distribution factorize into the product of pure (Gaussian) diffusions $q_{x}(x_{0:T})$ and pure telegraph processes $q_{\mu}(\mu_{0:T})$. For the hybrid FFL model this means that
\begin{equation}
q(x_{M\,0:T},x_{S\,0:T},x_{T\,0:T},\mu_{M\,0:T},\mu_{S\,0:T},\mu_{T\,0:T}) = \prod_{i = [M,S,T]} q_{x_{i}}(x_{{i}\,0:T})q_{\mu_{i}}(\mu_{{i}\,0:T})
\end{equation}
where $x_{i\,0:T}$ and $\mu_{\,0:T}$ represent continuous-time sample paths for the processes $x_i$ and $\mu_i$, respectively, in the interval $[0,T]$. In addition, we assume that the switching rates of the posterior telegraph processes $q_{\mu_{i}}(\mu_{{i}\,0:T})$ do not dependent on the state of the correspondent upstream proteins. By means of these assumptions, we can rewrite the KL divergence in a sum of terms which can be minimised with an iterative procedure~\cite{Opper:approximate10,Ocone:reconstructing11}. Parameter estimation is performed in the same algorithm, through minimisation of the KL divergence with respect to the kinetics parameters. Note that to avoid the protein concentrations $x_i$ to become negative, we can set a constraints on the parameter values (e.g. $b>0$ for activation and $A+b>0$ for repression).

\begin{figure}
\begin{center}
\includegraphics[width=11cm]{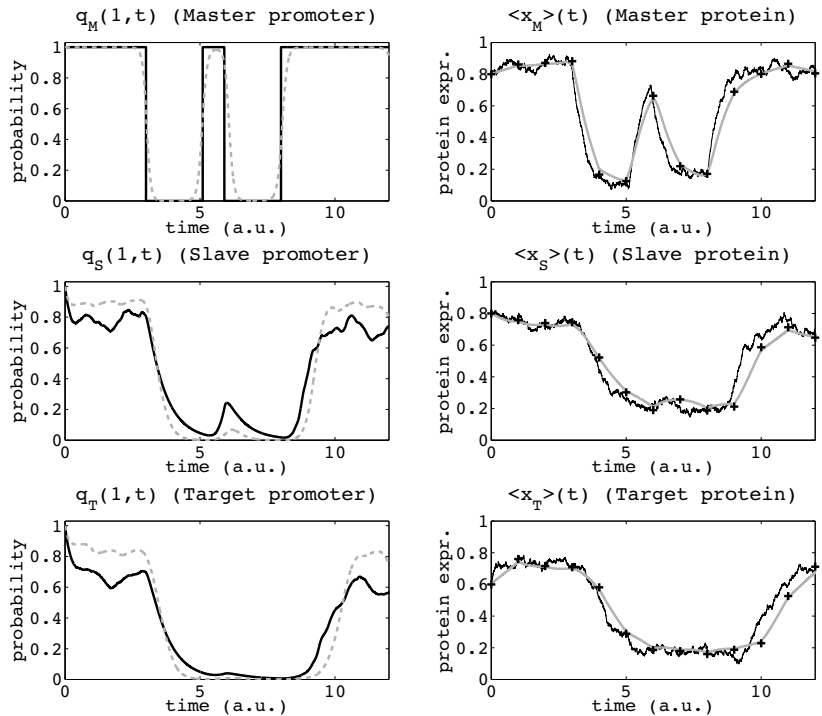}
\caption{Left: posterior promoter activity (dashed grey) compared to true simulated activity (solid black). Right: posterior protein states (solid grey) compared to noisy observations (black crosses). We also report simulated sample paths (solid black).}\label{fig003}
\end{center}
\end{figure}

\section{Results on a simulated data set}

We report a study of our hybrid FFL model on a simulated data set. Data are simulated using Equations~(\ref{FFL1})-(\ref{FFLend}) using the input function for the master promoter as showed in Figure~\ref{fig003} (top left corner, solid line). The master promoter has two main transitions: it transits from state $1$ to state $0$ around time $t=3$ and then back to state $1$ around time $t=8$. But more interestingly, we simulate a perturbation that switch its activity from $0$ to $1$ for a short time (at time $t=5$).

Protein noisy observations ($D_M$, $D_S$, $D_T$) are obtained by adding a zero mean Gaussian noise with standard deviation $0.01$ to the simulated true time courses ($x_M$, $x_S$, $x_T$).

Figure~\ref{fig003} shows results of our hybrid FFL model. The flexibility given by the latent promoter values provides posterior reconstructions of the protein states that fit well the noisy observations (Fig.~\ref{fig003}, right). More importantly, the model provides a qualitatively good reconstruction of the promoter states (Fig.~\ref{fig003}, left), especially during the transition times from one state to the other. In particular, it can capture the presence of the perturbation in the master promoter activity (Fig.~\ref{fig003}, top left corner) and its propagation to the slave promoter activity (Fig.~\ref{fig003}, center left). The perturbation is finally filtered out in the target promoter activity, which is correctly inferred from the model as well (Fig.~\ref{fig003}, bottom left).

\section{Discussion and conclusion}
Feedforward loop structures are ubiquitous regulatory motifs in biology due to their important signal processing functions. They are highly over-represented not only in microbial regulatory network, but in a variety of other contexts: for example, feed forward loops are an important structure in neuronal networks in the brain \cite{Sterratt:principles11}. In this paper, we present a statistical model of transcriptional FFLs based on the general framework for statistical modelling of regulatory networks of~\cite{OconeMillarSanguinetti}. We showed that the model has good identifiability, and indeed performs the filtering of fast transient information which is associated with real FFL networks. Compared to earlier attempts to model statistically FFLs \cite{Ocone:reconstructing11}, this paper uses a stochastic model of the system, and elegantly bypasses some of the most serious computational problems introduced by the saturating behaviour of the slave TF.

An important aspect which we have not touched upon in the present work is how the master and slave TF interact at the target promoter. Indeed, it is assumed that the contributions of the two TFs to the switching rates of the target promoters combine additively. Exploring different logics (e.g. OR, AND or XOR gates) remains an interesting development for future work.

\section*{Acknowledgements} This work was supported by an ERC starting independent research award to G.S. (grant reference ERC-306999-MLCS).

\nocite{*}
\bibliographystyle{eptcs}
\bibliography{HASbiblio}
\end{document}